\documentclass[runningheads]{llncs}
\usepackage{graphicx}

\usepackage{tikz}
\usepackage{comment}
\usepackage{amsmath,amssymb} 
\usepackage{color}
\usepackage{epsfig}
\usepackage{graphicx}
\usepackage{textcomp}
\usepackage{gensymb}
\usepackage{booktabs}
\usepackage{makecell}
\usepackage{enumitem}
\usepackage[ruled,vlined]{algorithm2e}
\usepackage[normalem]{ulem}
\usepackage{hyperref}
\usepackage[capitalise]{cleveref}
\usepackage{multirow}
\usepackage{wrapfig}

\usepackage[accsupp]{axessibility}  

\SetCommentSty{mycommfont}

\newcommand\boldred[1]{\textcolor{black}{\textbf{#1}}}

\newcommand{\nerfhist}{histogram-assisted NeRF\xspace}

\begin{document}
\pagestyle{headings}
\mainmatter
\def\ECCVSubNumber{4115}  

\title{DFNet: Enhance Absolute Pose Regression with Direct Feature Matching} 

\titlerunning{DFNet}

\author{Shuai Chen \and
Xinghui Li \and
Zirui Wang \and Victor A. Prisacariu\index{Prisacariu, Victor A.}}
\authorrunning{S. Chen et al.}

\institute{Active Vision Lab, University of Oxford}





\maketitle
\begin{abstract}

We introduce a camera relocalization pipeline that combines absolute pose regression (APR) and direct feature matching. By incorporating exposure-adaptive novel view synthesis, our method successfully addresses photometric distortions in outdoor environments that existing photometric-based methods fail to handle. With domain-invariant feature matching, our solution improves pose regression accuracy using semi-supervised learning on unlabeled data. In particular, the pipeline consists of two components: Novel View Synthesizer and DFNet. The former synthesizes novel views compensating for changes in exposure and the latter regresses camera poses and extracts robust features that close the domain gap between real images and synthetic ones. Furthermore, we introduce an online synthetic data generation scheme. We show that these approaches effectively enhance camera pose estimation both in indoor and outdoor scenes. Hence, our method achieves a state-of-the-art accuracy by outperforming existing single-image APR methods by as much as 56\%, comparable to 3D structure-based methods.\footnote{The code is available in \href{https://code.active.vision}{https://code.active.vision}.}
\keywords{Absolute Pose Regression, Feature Matching, NeRF}
\end{abstract}


\section{Introduction}
Estimating the position and orientation of cameras from images is essential in many applications, including virtual reality, augmented reality, and autonomous driving. While the problem can be approached via a geometric pipeline consisting of image retrieval, feature extraction and matching, and a robust Perspective-n-Points (PnP) algorithm, many challenges remain, such as invariance to appearance or the selection of the best set of method hyperparameters.

Learning-based methods have been used in traditional pipelines to improve robustness and accuracy, e.g. by generating neural network (NN)-based feature descriptors \cite{DeTone18,Li20,Lindenberger21,Sarlin20,sarlin21pixloc}, combining feature extraction and matching into one network \cite{Shotton13}, or incorporating differentiable outlier filtering modules \cite{Brachmann17,Brachmann18,brachmann2020dsacstar}. Although deep 3D-based solutions have demonstrated favorable results, many pre-requisites often remain, such as the need for an accurate 3D model of the scene and manual hyperparameter tuning of the remaining classical components.

The alternative end-to-end NN-based approach, termed absolute pose regression (APR), directly regresses the absolute pose of the camera from input images \cite{Kendall15} without requiring prior knowledge about the 3D structure of the neighboring environment. Compared with deep 3D-based methods, APR methods can achieve at least one magnitude faster running speeds at the cost of inferior accuracy and longer training time. Although follow-up works such as MapNet \cite{Brahmbhatt18} and Kendall \textit{et al.} \cite{Kendall17} attempt to improve APR methods by adding various constraints such as relative pose and scene geometry reprojection, a noticeable gap remains between APR and 3D-based methods.

Recently, Direct-PN \cite{chen21} achieved state-of-the-art (SOTA) accuracy in indoor localization tasks among existing single-frame APR methods. As well as being supervised by ground-truth poses, the network directly matches the input image and a NeRF-rendered image at the predicted pose. However, it has two major limitations: (a) direct matching is very sensitive to photometric inconsistency, as images with different exposures could produce a high photometric error even from the same camera pose, which reduces the viability of photometric direct matching in environments with large photometric distortions, such as outdoor scenes; (b) there is a domain gap between real and rendered images caused by poor rendering quality or changes in content and appearance of the query scene. 

In order to address these limitations, we propose a novel relocalization pipeline that combines APR and direct feature matching. First, we introduce a histogram-assisted variant of NeRF, which learns to control synthetic appearance via histograms of luminance information. This significantly reduces the gap between real and synthetic image appearance. Second, we propose a network \textit{DFNet} that extracts domain invariant features and regresses camera poses, trained using a contrastive loss with a customized mining method. Matching these features instead of direct pixels colors boosts the performance of the direct dense matching further. Third, we improve generalizability by (i) applying a cheap Random View Synthesis (RVS) strategy to efficiently generate a synthetic training set by rendering novel views from randomly generated pseudo training poses and (ii) allow the use of unlabeled data. We show that our method outperforms existing single-frame APR methods by as much as 56\% on both indoor 7-Scenes and outdoor Cambridge datasets. We summarize our main contributions as follows:

\begin{enumerate}[leftmargin=*]
    \item We introduce a direct feature matching method that offers better robustness than the prior photometric matching formulation, and devise a network DFNet that can effectively bridge the feature-level domain gap between real and synthetic images.

    \item We introduce a histogram-assisted NeRF, which can scale the direct matching approach to scenes with large photometric distortions, \textit{e.g.}, outdoor environments, and provide more accurate rendering appearance to unseen real data.
    
    \item We show that a simpler synthetic data generation strategy such as RVS can improve pose regression performance.
\end{enumerate}

\begin{figure*}[t]
   \begin{center}
   \includegraphics[width=0.8\textwidth]{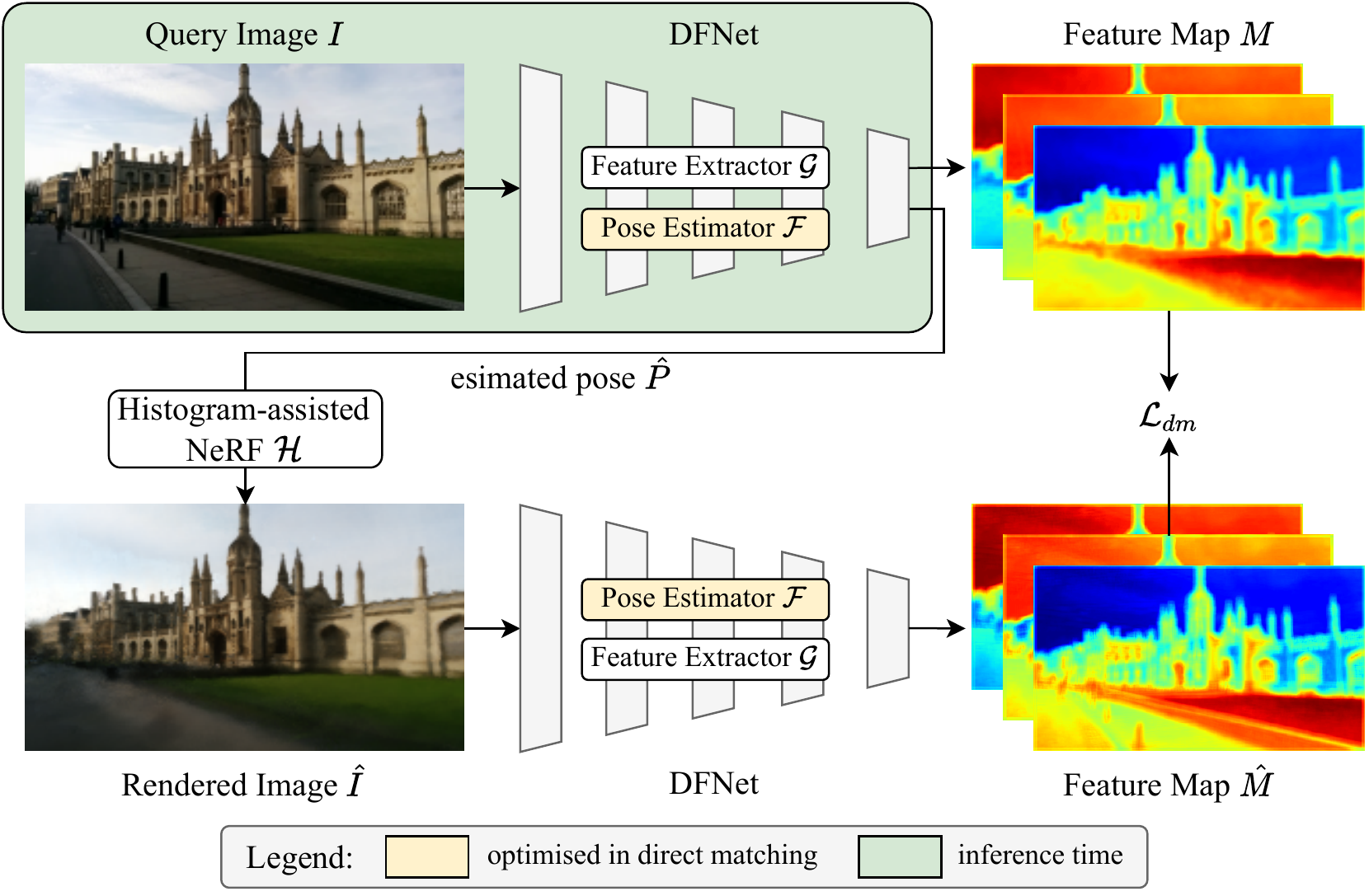}
   \end{center}

   \caption{\textbf{Overview of the direct feature matching pipeline}. Given an input image $I$, a pose regressor $\mathcal{F}$ estimates a camera pose $\hat{P}$, from which a luminance prior NVS system $\mathcal{H}$ renders a synthetic image $\hat{I}$. Domain invariant features of $M$ and $\hat{M}$ are extracted using a feature extractor $\mathcal{G}$, supplying a feature-metric direct matching signal $\mathcal{L}_{dm}$ to optimize the pose regressor.}
   \label{fig:DFNet} 
\end{figure*}
\section{Related Work} \label{sec:related_work}
\textbf{Absolute Pose Regression} Absolute pose regression aims to directly regress the 6-DOF camera pose from an image using Convolutional Neural Networks. The first practice in this area is introduced by PoseNet~\cite{Kendall15}, which is a GoogLeNet-backbone network appended with an MLP regressor. Successors of PoseNet propose several variations in network architectures, such as adding LSTM layers~\cite{Walch17}, adapting an encoder-decoder backbone~\cite{Melekhov17}, splitting the network into position and orientation branches~\cite{Wu17}, or incorporating attentions using transformers~\cite{Shavit21,Shavit21multiscene}. Other methods propose different strategies to train APR. Bayesian PoseNet \cite{Kendall16Bayesian} inserts Monte Carlo dropout to a Bayesian CNN that estimates pose with uncertainty. Kendall \textit{et al.}~\cite{Kendall17} proposes to balance the translation and rotation loss at training using learnable weights and reprojection error. MapNet~\cite{Brahmbhatt18} trains the network using both absolute pose loss and relative pose loss but can infer in a single-frame manner. Direct-PoseNet (Direct-PN)~\cite{chen21} adapts additional photometric loss by comparing the query image with NeRF synthesis on the predicted pose.

\noindent\textbf{Semi-supervised Learning in APR} Several APR methods explore semi-supervised learning with additional images without ground-truth pose annotation to improve pose regression performance. To the best of our knowledge, MapNet+~\cite{Brahmbhatt18} and MapNet+PGO~\cite{Brahmbhatt18} are the pioneers to train APR on unlabeled video sequences using external VO algorithms \cite{Engel17,Engel13}. Direct-PN+ \cite{chen21} finetune on unlabeled data from arbitrary viewpoints solely based on its direct matching formulation. While the direct matching idea from Direct-PN+ inspires our proposed method, we focus on training in the feature space. Our solution can scale to scenes with large photometric distortion, where the previous method fails.

\noindent\textbf{Novel View Synthesis in APR} Novel View Synthesis (NVS) can be beneficial to the visual relocalization task. For example, NVS can expand training space by generating extra synthetic data. Purkait \textit{et al.} \cite{Purkait18} propose a method to generate realistic synthetic training data for pose regression leveraging the 3D map and feature correspondences. LENS \cite{Moreau21} deploys a NeRF-W \cite{martinbrualla2020nerfw} model to sample the scene boundaries and synthesize virtual views with uniformly generated virtual camera poses. However, Purkait \textit{et al}. rely on a pre-computed reconstructed 3D map. LENS is limited by its costly offline computation efficiency and the lack of compensation to the domain gap between synthetic and real images, i.e., dynamic objects or artifacts. Another direction is to embed NVS into the pose estimation process. InLoc \cite{Taira18} verifies the predicted pose with view synthesis. Ng et al. \cite{Ng21} combine a multi-view stereo (MVS) model with a relative pose regressor (RPR). iNeRF \cite{yen2020inerf}, Wang \textit{et al.} \cite{wang2021nerfmm}, and Direct-PN \cite{chen21} utilize an inverted NeRF to optimize the camera pose. Our paper is the first to incorporate both strategies yet have major differences from the above methods. 1) we introduce an NVS method that can adapt to real exposure change in view synthesis. 2) we address the domain adaptation problem between the actual camera footage with synthetic images. 3) our synthetic data generation strategy is comparatively less constrained and can be deployed efficiently in online training.

\section{Method} \label{sec:method}
We illustrate our proposed direct feature matching pipeline in \cref{fig:DFNet}, which contains two primary components: 
1) the DFNet network, which, given an input image $I$, uses a pose estimator $\mathcal{F}$ to predict a 6-DoF camera pose and a feature extractor $\mathcal{G}$ to compute a feature map $M$, and 
2) a \nerfhist $\mathcal{H}$, which compensates for high exposure fluctuation by providing luminance control when rendering a novel view given an arbitrary pose.

Training the direct feature matching pipeline can be split into two stages, (i) DFNet and the \nerfhist, and (ii) direct feature matching.
In stage one, we train the NVS module $\mathcal{H}$ like a standard NeRF, and the DFNet with a loss term $\mathcal{L}_{DFNet}$ in \cref{eq:L_fnet}.
In stage two, fixing the \nerfhist and the feature extractor $\mathcal{G}$, we further optimize the main pose estimation module $\mathcal{F}$ via a direct feature matching signal between feature maps extracted from the real image and its synthetic counterpart $\hat{I}$, which is rendered from the predicted pose $\hat{P}$ of image $I$ via the NVS module $\mathcal{H}$. At test time, only the pose estimator $\mathcal{F}$ is required given the query image, which ensures a rapid inference.

This section is organized as follows: the DFNet pipeline is detailed in \cref{sec:direct_feat_match_for_pose}, followed by a showcase of our \nerfhist $\mathcal{H}$ in \cref{sec:histnerf}. 
To further boost the pose estimation accuracy, an efficient Random View Synthesis (RVS) training strategy is introduced in \ref{sec:rvs}.

\subsection{Direct Feature Matching For Pose Estimation}\label{sec:direct_feat_match_for_pose}
\begin{figure*}[t]
   \begin{center}
  \includegraphics[width=\textwidth]{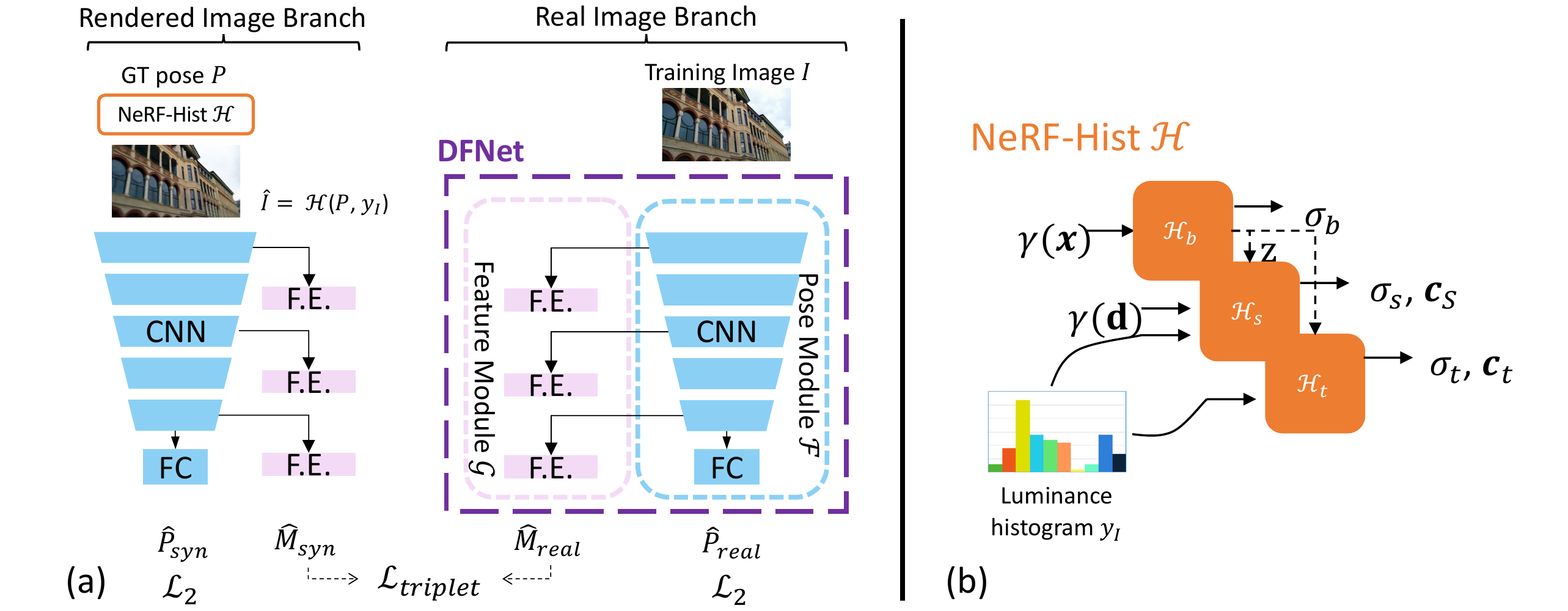}
   \end{center}
   \caption{(a) The training scheme for DFNet to close the domain gap between real images and rendered images. (b) The \nerfhist architecture.}
   \label{fig:FNet}
\end{figure*}
This section aims to introduce:
1) the design of our main network DFNet, 
2) the direct feature matching formulation that boosts pose estimation performance in a semi-supervised training manner, and
3) the contrastive-training scheme that closes the domain gap between real images and synthetic images.

\subsubsection{DFNet Structure}
The DFNet in our pipeline consists of two networks, a pose estimator $\mathcal{F}$ and a feature extractor $\mathcal{G}$. The pose estimator $\mathcal{F}$ in our DFNet is similar to an ordinary PoseNet, which predicts a 6-DoF camera pose $\hat{P} = \mathcal{F}(I)$ for an input image $I$, and can be supervised by an $L_1$ or $L_2$ loss between the pose estimation $\hat{P}$ and its ground truth pose $P$.

The feature extractor $\mathcal{G}$ in our DFNet takes as input feature maps extracted from various convolutional blocks in the pose estimator and pushes them through a few convolutional blocks, producing the final feature maps $M =\mathcal{G}(I)$, which are the key ingredients during feature-metric direct matching.

Two key properties of the feature extractor $\mathcal{G}$ that we seek to learn are 1) domain invariance, i.e., being invariant to the domain of real images and the domain of synthetic images and 
2) transformation sensitive, i.e., being sensitive to the image difference that is caused by geometry transformations. 
With these properties learned, our feature extractor can extract domain-invariant features during feature-metric direct matching while preserving geometry-sensitive information for pose learning. 
We detail the way to train the DFNet in the \emph{Closing the Domain Gap} section.

\subsubsection{Direct Feature Matching}
\label{sec:feat_metric_dm}
Direct matching in APR was first introduced by Direct-PN \cite{chen21}, which minimizes the photometric difference between a real image $I$ and a synthetic image $\hat{I}$ rendered from the estimated pose $\hat{P}$ of the real image $I$. 
Ideally, if the predicted pose $\hat{P}$ is close to its ground truth pose $P$, and the novel view renderer produces realistic images, the rendered image $\hat{I}$ should be indistinguishable from the real image.

In practice, we found the photometric-based supervision signal could be noisy in direct matching, when part of scene content changes. For example, random cars and pedestrians may appear through time or the NeRF rendering quality is imperfect.
Therefore, we propose to measure the distance between images in feature space instead of in photometric space, given that the deep features are usually more robust to appearance changes and imperfect renderings.

Specifically, for an input image $I$ and its pose estimation $\hat{P}=\mathcal{F}(I)$, a synthetic image $\hat{I} = \mathcal{H}(\hat{P}, \mathbf{y}_I)$ can be rendered using the pose estimation $\hat{P}$ and the histogram embedding $\mathbf{y}_I$ of the input image $I$. 
We then extract the feature map $M \in \mathbb{R}^{H_M \times W_M \times C_M}$ and $\tilde{M} \in \mathbb{R}^{H_M \times W_M \times C_M}$ for image $I$ and $\hat{I}$ respectively, where $H_M$ and $W_M$ are the spatial dimensions and $C_M$ is the channel dimension of the feature maps. 
To measure the difference between two feature maps, we compute a cosine similarity between feature $m_i \in \mathbb{R}^{C_M}$ and $\tilde{m}_i \in\mathbb{R}^{C_M}$ for each feature location $i$:
\begin{equation}
\cos(m_i, \tilde{m}_i) = \frac{m_i\cdot\tilde{m}_i} {\|m_{i}\|_{2}\cdot\|\tilde{m}_{i}\|_{2}}.
\end{equation}
By minimizing the feature-metric direct matching loss $\mathcal{L}_{dm} = \sum_i ( 1-\cos(m_i, \tilde{m}_i) )$, the pose estimator $\mathcal{F}$ can be trained in a semi-supervised manner (note no ground truth label required for the input image $I$). 

Our direct feature matching may optionally follow the procedure of semi-supervised training proposed by MapNet+ \cite{Brahmbhatt18} to improve pose estimation with unlabeled sequences captured in the same scene. Unlike \cite{Brahmbhatt18}, which requires sequential frames to enforce a relative geometric constraint using a VO algorithm, our feature-matching can be trained by images from arbitrary viewpoints without ground truth pose annotation. Our method can be used at train time with a batch of unlabeled images, or as a pose refiner for a single test image. In the latter case, our direct matching can also be regarded as a post-processing module. During the training stage, only the weights of the pose estimator will be updated, whereas the feature extractor part remains frozen to back-propagation.

\subsubsection{Closing the Domain Gap}
\begin{figure*}[t]
   \begin{center}
   \includegraphics[width=0.8\textwidth]{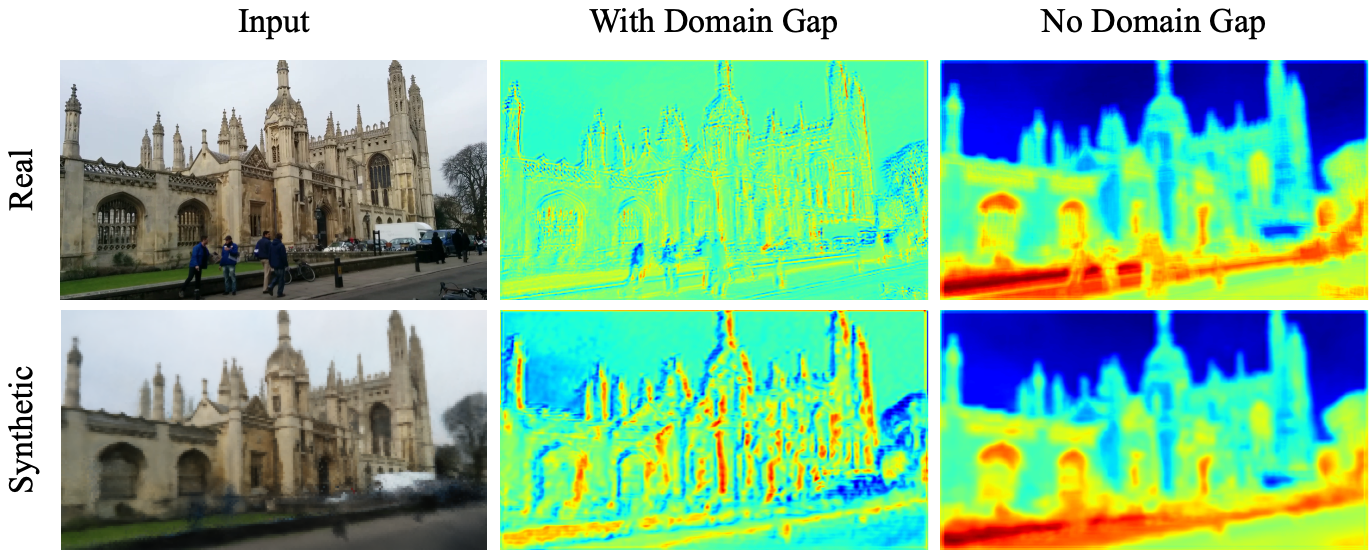}
   \end{center}
\caption{A visual comparison of features before and after closing the domain gap. Ideally, a robust feature extractor shall produce indistinguishable features between real and rendered images from the same pose. Column 2/Column 3 are features trained without/with using our proposed $\mathcal{L}_{triplet}$ loss, where our method can effectively produce similar features across two domains.}
   \label{fig:domain_gap}
\end{figure*}
We notice that synthetic images from NeRF are imperfect due to rendering artifacts or lack of adaption of the dynamic content of the scene, which leads to a domain gap between render and real images.
This domain gap poses difficulties to our feature extractor (\cref{fig:domain_gap}), which we expect to produce features far away if two views are from different poses and to produce similar features between a rendered view and a real image from the same pose.

Intuitively, we could simply enforce the feature extractor to produce similar features for a rendered image $\hat{I}$ and a real image $I$ via a distance function $d(\cdot)$ during training.
However, this approach leads to model collapse \cite{Chen21explore}, which motivates us to explore the original triplet loss:
\begin{equation}
    \mathcal{L}_{triplet}^{ori}=\max \left\{d(M^{P}_{real}, M^{P}_{syn}) -  d(M^{P}_{real}, M^{\bar{P}}_{syn}) +\operatorname{margin}, 0\right\},
    \label{eq:ori_triplet}
\end{equation}
where $M^{P}_{real}$ and $M^{P}_{syn}$, the feature maps of a real image and a synthetic image at pose $P$, compose a positive pair, and $M^{\bar{P}}_{syn}$ is a feature map of a synthetic image rendered at an arbitrary pose $\bar{P}$ other than the pose $P$.

With a closer look at the task of feature-metric direct matching, we implement a customized in-triplet mining which explores the minimum distances among negative pairs:
\begin{equation}
    \mathcal{L}_{triplet}=\max \left\{d(M^{P}_{real}, M^{P}_{syn}) - q_{\ominus} + \operatorname{margin}, 0\right\},
\end{equation}
where the positive pair is as same as \cref{eq:ori_triplet} and $q_{\ominus}$ is the minimum distance between four negative pairs:
\begin{equation}
    q_{\ominus} = \min \left\{ 
    d(M^{P}_{real}, M^{\bar{P}}_{real}),
    d(M^{P}_{real}, M^{\bar{P}}_{syn}),
    d(M^{P}_{syn}, M^{\bar{P}}_{real}),
    d(M^{P}_{syn}, M^{\bar{P}}_{syn})
    \right\},
\end{equation}
which essentially takes the hardest negative pair among all matching pairs between synthetic images and real images that are in different camera poses. 
The margin value is set to $1.0$ in our implementation. Since finding the minimum of negative pairs is non-differentiable, we implement the in-triplet mining as a prior step before $\mathcal{L}_{triplet}$ is computed.

Overall, to train the pose estimator and to obtain domain invariant and transformation sensitive property, we adapt a siamese-style training scheme as illustrated in \cref{fig:FNet}a.
Given an input image $I$ and its ground truth pose $P$, a synthetic image $\hat{I}$ can be rendered via the NVS module $\mathcal{H}$ (assumed pre-trained) using the ground truth pose $P$. We then present both the real image $I$ and the synthetic image $\hat{I}$ to the pose estimator and the feature extractor, resulting in pose estimations $\hat{P}_{real}$ and $\hat{P}_{syn}$ and feature maps $M_{real}$ and $M_{syn}$ for the real image $I$ and synthetic image $\hat{I}$, respectively. The training then is supervised via a combined loss function
\begin{equation}
    \mathcal{L}_{DFNet} = \mathcal{L}_{triplet} + \mathcal{L}_{RVS} +  \frac{1}{2}(\|P-\hat{P}_{real}\|_{2} + \|P-\hat{P}_{syn}\|_{2}), 
    \label{eq:L_fnet}
\end{equation}
where $\| \cdot \|$ denotes a $L_2$ loss and $\mathcal{L}_{RVS}$ is a supervision signal from our RVS training strategy, which we explain in \cref{sec:rvs}.

\subsection{Histogram-assisted NeRF}\label{sec:histnerf}

\begin{figure}[t]
  \centering
    \begin{tabular}{c}
      \includegraphics[width=0.95\linewidth]{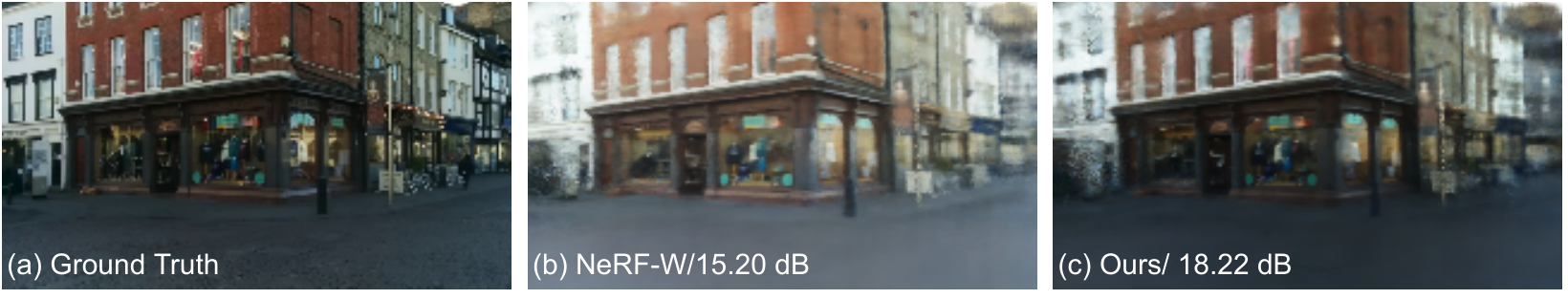}
    \end{tabular}
\caption{Typically NeRF only renders views that reflect the appearance of its training sequences, as shown by NeRF-W's synthetic view (b). However, in relocalization tasks, the query set may have different appearances or exposures to the train set. The proposed histogram-assisted NeRF (c) can render a more accurate appearance to the unseen query set (a) in both quantitative (PSNR) and visual comparisons. We refer to the supplementary for more examples.}

\label{fig:nerf-hist}
\end{figure}
The DFNet pipeline relies on an NVS module that renders a synthetic image from which we extract a feature map and compare it with a real image. 
Theoretically, while the NVS module in our pipeline can be in any form as long as it provides high-quality novel view renderings, 
in practice, we found that due to the presence of auto exposure during image capturing, it is necessary to have a renderer that can render images in a compensated exposure condition. 
Although employing direct matching in feature space could mediate the exposure issue to some extent, we find decoupling the exposure issue from the domain adaption issue leads to better pose estimation results.

One off-the-shelf option is a recent work NeRF-W \cite{martinbrualla2020nerfw}, which offers the ability to control rendered appearance via an appearance embedding that is based on frame indices. 
However, in the context of direct matching, since we aim to compare a real image with its synthetic version, we desire a more fine-grained exposure control to render an image that matches the exposure condition of the real image, as illustrated in \cref{fig:nerf-hist}.

To this end, we propose a novel view renderer \nerfhist (\cref{fig:FNet}b) which renders an image $\hat{I} = \mathcal{H}(P, \mathbf{y}_I)$ that matches the exposure level of a query real image $I$ via a histogram embedding $\mathbf{y}_I$ of the query image $I$ at an arbitrary camera pose $P$. 
Specifically, our NeRF contains 3 components:
\begin{enumerate}
    \item 
    A base network $\mathcal{H}_{b}$ that provides a density estimation $\sigma_b$ and a hidden state $\mathbf{z}$ for a coarse estimation:
    $
        [\sigma_b, \mathbf{z}]=\mathcal{H}_{b}(\gamma(\mathbf{x})).
    $
    \item
    A static network $\mathcal{H}_{s}$ to model density $\sigma_s$ and radiance $\mathbf{c}_s$ for static structure and appearance:
    $
        [\sigma_s, \mathbf{c}_s]=\mathcal{H}_{s}(\mathbf{z}, \gamma(\mathbf{d}), \mathbf{y}_I).
    $
    \item
    A transient network $\mathcal{H}_{t}$ to model density $\sigma_t$, radiance $\mathbf{c}_t$ and an uncertainty estimation $\beta$ for dynamic objects: 
    $
        [\sigma_t, \mathbf{c}_t, \beta]=\mathcal{H}_{t}(\mathbf{z}, \mathbf{y}_I).
    $
\end{enumerate}
As for the input, $\mathbf{x}$ is a 3D point and $\mathbf{d}$ is a view angle that observes the 3D point, with both of them encoded by a positional encoding \cite{gehring2017convolutional,vaswani2017attention,Mildenhall20} operator $\gamma(\cdot)$ before injecting to each network.

During training, the coarse density estimation from the base network $\mathcal{H}_{b}$ provides a distribution where the other two networks could sample more 3D points near non-empty space accordingly. 
Both the static and the transient network are conditioned on a histogram-based embedding $\mathbf{y}_I \in \mathbb{R}^{C_y}$, which is mapped from a $N_b$ bins histogram.
The histogram is computed on the luma channel Y of a target image in YUV space. We found this approach works well in a direct matching context, not only in feature-metric space but also in photometric space.

We adopt a similar network structure and volumetric rendering method as in NeRF-W\cite{martinbrualla2020nerfw}, to which we refer readers for more details.

\subsection{Random View Synthesis}\label{sec:rvs}
During the training of DFNet, we can generate training data by synthesis more views from randomly perturbed training poses. We refer this process as Random View Synthesis (RVS), and we use this data generation strategy to help the DFNet to better generalize to unseen views. 

Specifically, given a training pose $P$, a perturbed pose $P^\prime$ can be generated around the training pose with a random translation noise of $\psi$ meters and random rotation noise of $\phi$ degrees. 
A synthetic image $I^\prime = \mathcal{H}(P^\prime, \mathbf{y}_{I_{nn}})$ is then rendered via \nerfhist 
$\mathcal{H}$, with $\mathbf{y}_{I_{nn}}$ being the histogram embedding of the training image with the nearest training pose. 
The synthetic pose-image pair $(P^\prime, I^\prime)$ is used as a training sample for the pose estimator to provide an additional supervision signal $\mathcal{L}_{RVS} = \|P^\prime - \hat{P^{\prime}}\|_2$, where $\hat{P^{\prime}} = \mathcal{F}(I^\prime)$ is the pose estimation of the rendered image.

A key advantage of our method is efficiency in comparison with prior training sample generation methods. For example, LENS \cite{Moreau21} generates high-resolution synthetic data with a maximum of 40s/image and requires complicated parameter settings in finding candidate poses within scene volumes. In contrast, our RVS is a lightweight strategy that seamlessly fits our DFNet training at a much cheaper cost (12.2 fps) and with fewer constraints in pose generation while being able to reach similar performance. We refer to \cref{sec:effectiveness_of_rvs} for more discussion.

\section{Experiments} \label{sec:experiments}

\begin{table}[t]
\caption{\textbf{Pose regression results on 7-Scenes dataset.} We compare DFNet and DFNet$_{dm}$ (DFNet with feature-metric direct matching) with prior single-frame APR methods and unlabeled training methods, in median translation error (m) and rotation error (\degree). Note that MapNet+ and MapNet+PGO are sequential methods with unlabeled training. Numbers in \boldred{bold} represent the best performance.}
\label{table:1}
\resizebox{0.99\textwidth}{!}{
\begin{tabular}{c|l|ccccccccccccc|c}
\toprule
& Methods & Chess && Fire && Heads && Ofﬁce && Pumpkin && Kitchen && Stairs & Average \\

\midrule
\multirow{11}{*}{\makecell[c]{1-frame\\APR}}
& PoseNet(PN)\cite{Kendall15}         & 0.32/8.12   && 0.47/14.4   && 0.29/12.0   && 0.48/7.68   && 0.47/8.42   && 0.59/8.64   && 0.47/13.8 & 0.44/10.4 \\
& PN Learn $\sigma^2$\cite{Kendall17}   & 0.14/4.50   && 0.27/11.8   && 0.18/12.1   && 0.20/5.77   && 0.25/4.82   && 0.24/5.52   && 0.37/10.6   & 0.24/7.87 \\
& geo. PN\cite{Kendall17}              & 0.13/4.48   && 0.27/11.3   && 0.17/13.0   && 0.19/5.55   && 0.26/4.75   && 0.23/5.35   && 0.35/12.4   & 0.23/8.12 \\
& LSTM PN\cite{Walch17}              & 0.24/5.77   && 0.34/11.9   && 0.21/13.7   && 0.30/8.08   && 0.33/7.00   && 0.37/8.83   && 0.40/13.7   & 0.31/9.85 \\
& Hourglass PN\cite{Melekhov17}         & 0.15/6.17   && 0.27/10.8  && 0.19/11.6   && 0.21/8.48   && 0.25/7.0    && 0.27/10.2   && 0.29/12.5   & 0.23/9.53 \\
& BranchNet\cite{Wu17}            & 0.18/5.17   && 0.34/8.99   && 0.20/14.2   && 0.30/7.05   && 0.27/5.10   && 0.33/7.40   && 0.38/10.3   & 0.29/8.30 \\
& MapNet\cite{Brahmbhatt18}               & 0.08/3.25   && 0.27/11.7   && 0.18/13.3   && 0.17/5.15   && 0.22/4.02   && 0.23/\boldred{4.93}   && 0.30/12.1   & 0.21/7.77 \\
& Direct-PN\cite{chen21}            & 0.10/3.52   && 0.27/8.66   && 0.17/13.1   && 0.16/5.96   && 0.19/3.85   && 0.22/5.13   && 0.32/10.6  & 0.20/7.26 \\
& TransPoseNet\cite{Shavit21}         & 0.08/5.68   && 0.24/10.6   && 0.13/12.7   && 0.17/6.34   && 0.17/5.6    && 0.19/6.75   && 0.30/7.02   & 0.18/7.78 \\
& MS-Transformer\cite{Shavit21multiscene}       & 0.11/4.66   && 0.24/9.60  && 0.14/12.2  && 0.17/5.66   && 0.18/4.44   && \boldred{0.17}/5.94   && 0.17/5.94   & 0.18/7.28 \\
& DFNet (ours)         & \boldred{0.05}/\boldred{1.88}   && \boldred{0.17}/\boldred{6.45}   && \boldred{0.06}/\boldred{3.63}   && \boldred{0.08}/\boldred{2.48}   && \boldred{0.10}/\boldred{2.78}   && 0.22/5.45   && \boldred{0.16}/\boldred{3.29}   & \boldred{0.12}/\boldred{3.71} \\

\midrule
\multirow{4}{*}{\makecell[c]{Unlabel\\Data}}
& MapNet$_{+(seq.)}$\cite{Brahmbhatt18}       & 0.10/3.17   && 0.20/9.04   && 0.13/11.1   && 0.18/5.38   && 0.19/3.92   && 0.20/5.01   && 0.30/13.4   & 0.19/7.29 \\
& MapNet$_{+PGO(seq.)}$\cite{Brahmbhatt18}    & 0.09/3.24   && 0.20/9.29   && 0.12/8.45   && 0.19/5.42   && 0.19/3.96   && 0.20/4.94   && 0.27/10.6   & 0.18/6.55 \\
& Direct-PN+U\cite{chen21} & 0.09/2.77   && 0.16/4.87   && 0.10/6.64   && 0.17/5.04   && 0.19/3.59   && 0.19/4.79   && 0.24/8.52   & 0.16/5.17  \\
& DFNet$_{dm}$ (ours)        & \boldred{0.04}/\boldred{1.48}   && \boldred{0.04}/\boldred{2.16}   && \boldred{0.03}/\boldred{1.82}  && \boldred{0.07}/\boldred{2.01}   && \boldred{0.09}/\boldred{2.26}   && \boldred{0.09}/\boldred{2.42}   && \boldred{0.14}/\boldred{3.31}   & \boldred{0.07}/\boldred{2.21}  \\

\bottomrule
\end{tabular}
}
\end{table}

\subsection{Implementation}
We introduce the implementation details for histogram-assisted NeRF, DFNet, and direct feature matching. We also provide more details in the supplementary.

\textbf{NeRF}
Our \nerfhist model is trained with a re-aligned and re-centered pose in SE(3), similar to Mildenhall \textit{et al.} \cite{Mildenhall20}. The image histogram bin size is set to $N_b=10$ and embedded with a vector dimension of 50 for the static model and 20 for the transient model. We train the model with a learning rate of $5 \times 10^{-4}$ and an exponential decay of $5 \times 10^{-4}$ for 600 epochs.

\textbf{DFNet}
Our DFNet adapts an ImageNet pre-trained VGG-16 \cite{Simonyan15} as the backbone, and an Adam optimizer with a learning rate of $1 \times 10^{-4}$ is applied during training.
For feature extraction, we extract $L=3$ feature maps from the end of the encoder's first, third, and fifth blocks before pooling layers. All final feature outputs are upscaled to the same size as the input image $H \times W$ with bilinear upsampling. 
For pose regression, we regresses the SE(3) camera pose with a fully connected layer. A singular value decomposition (SVD) is applied to ensure the rotation component of $\hat{P}$ is normalized \cite{chen21}.

\textbf{Direct Feature Matching}
To validate our feature-metric direct matching formulation, we follow the same procedure from MapNet+ \cite{Brahmbhatt18} and Direct-PN+U \cite{chen21}, which use a portion of validation images without the ground truth poses for finetuning. 
When finetuning DFNet, we optimize the pose regression module $\mathcal{F}$ solely based on the direct feature matching loss $\mathcal{L}_{dm}$. 
We set the batch size to 1 and the learning rate to $1 \times 10^{-5}$. 
For naming simplicity, we named our model trained with direct feature matching as DFNet$_{dm}$. 

\subsection{Evaluation on the 7-Scenes Dataset}
We evaluate our method on an indoor camera localization dataset 7-Scenes \cite{Glocker13,Shotton13}. The dataset consists of seven indoor scenes scaled from $1m^3$ to $18m^3$. Each scene contains 1000 to 7000 training sets and 1000 to 5000 validation sets. Both \nerfhist and DFNet use subsampled training data with a spacing window $d=5$ for scenes containing $\leq$ 2000 frames and $d=10$ otherwise.
RVS poses are sampled on the training pose, and the DFNet parameters are $t_\psi=0.2m$, $r_\phi=10\degree$, and $d_{max}=0.2m$.
For fair comparison to other unlabeled training methods such as MapNet+ and Direct-PN, we finetune our DFNet$_{dm}$ using the same amount of unlabeled samples, which is 1/5 or 1/10 of the sequences based on the spacing window above to ensure our method is not overfitting to the entire test sequences.

We compared our method quantitatively with prior single-frame APR methods and unlabeled training APR methods in \cref{table:1}. The results show that both our DFNet and DFNet$_{dm}$ obtain superior accuracy, and DFNet$_{dm}$ achieves 56\% and 57\% improvement over averaged median translation and rotation errors compared to prior SOTA performance.


\begin{table}[t]

\caption{\textbf{Single-frame APR results on Cambridge dataset.} We report the median position and orientation errors in $m/\degree$ and the respective rankings over scene average as in \cite{Shavit21,Shavit21multiscene}. The best results is highlighted in \boldred{bold}. For fair comparisons, we omit prior APR methods which did not publish results in Cambridge.}

\label{table:2}
\centering
\resizebox{0.87\textwidth}{!}{
\begin{tabular}{l|ccccccc|c|c|c}
\toprule
Methods            & Kings     && Hospital  && Shop      && Church    & Average  & Ranks & Final Rank \\
\midrule
PoseNet(PN)\cite{Kendall15}       & 1.66/4.86 && 2.62/4.90 && 1.41/7.18 && 2.45/7.96 & 2.04/6.23 & 9/9 & 9\\
PN Learn $\sigma^2$\cite{Kendall17} & 0.99/1.06 && 2.17/2.94 && 1.05/3.97 && 1.49/3.43 & 1.43/2.85 & 6/3 & 5\\
geo. PN\cite{Kendall17}            & 0.88/1.04 && 3.20/3.29 && 0.88/3.78 && 1.57/3.32 & 1.63/2.86 & 7/4 & 6\\
LSTM PN\cite{Walch17}            & 0.99/3.65 && 1.51/4.29 && 1.18/7.44 && 1.52/6.68 & 1.30/5.51 & 5/8 & 7\\
MapNet\cite{Brahmbhatt18}             & 1.07/1.89 && 1.94/3.91 && 1.49/4.22 && 2.00/4.53 & 1.63/3.64 & 7/7 & 8\\
TransPoseNet\cite{Shavit21}       & 0.60/2.43 && 1.45/3.08 && 0.55/3.49 && 1.09/4.94 & 0.91/3.50 & 2/6 & 3\\
MS-Transformer\cite{Shavit21multiscene} & 0.83/1.47 && 1.81/2.39 && 0.86/3.07 && 1.62/3.99 & 1.28/2.73 & 4/2 & 2\\
DFNet (ours)       & 0.73/2.37 && 2.00/2.98 && 0.67/2.21 && 1.37/4.03 & 1.19/2.90 & 3/5 & 3\\
DFNet$_{dm}$ (ours)      & \boldred{0.43}/\boldred{0.87} && \boldred{0.46}/\boldred{0.87} && \boldred{0.16}/\boldred{0.59} && \boldred{0.50}/\boldred{1.49} & \boldred{0.39}/\boldred{0.96} & \boldred{1}/\boldred{1} & \boldred{1}\\
\bottomrule
\end{tabular}
}
\end{table}

\subsection{Evaluation on Cambridge Dataset}
We further compare our approach on four outdoor scenes from the Cambridge Landmarks \cite{Kendall15} dataset, scaling from $875m^2$ to $5600m^2$. Each scene contains from 200+ to 1500 training samples. Our models are trained with 50\% of training data, and DFNet's RVS are $t_\psi=3m$, $r_\phi=7.5\degree$, and $d_{max}=1m$. For finetuning DFNet$_{dm}$ with unlabeled data, we use 50\% of the unlabeled validation sequence since fewer validation sets are available than 7-Scenes.
\cref{table:2} shows a comparison between our approach and prior single-frame APR methods, which omits prior APR methods that did not report results in Cambridge. We observe that our DFNet$_{dm}$ outperforms other methods significantly (60\%+ in scene average), which further proves the effectiveness of our approach.

\begin{table*}[t]
\caption{Comparison between our method and sequential-based APR methods and 3D structure-based methods.}
\label{table:3}
\centering
\resizebox{0.73\textwidth}{!}{
\begin{tabular}{l||c||cccc|c}
\toprule
          & 3D  & \multicolumn{4}{c}{Seq. APR} & 1-frame \\
\midrule
 Methods  & AS\cite{Sattler12} & \makecell[c]{MapNet\\+PGO\cite{Brahmbhatt18}} & CoordiNet\cite{Moreau21CoordiNet} & \makecell[c]{CoordiNet\\+Lens\cite{Moreau21}} & VLocNet\cite{Valada18} & DFNet$_{dm}$ \\
\midrule
 Chess    & 0.04/2.0      & 0.09/3.24  & 0.14/6.7  & 0.03/1.3       & 0.04/1.71 & 0.04/1.48        \\
 Fire     & 0.03/1.5      & 0.20/9.29  & 0.27/11.6 & 0.10/3.7       & 0.04/5.34 & 0.04/2.16        \\
 Heads    & 0.02/1.5      & 0.12/8.45  & 0.13/13.6 & 0.07/5.8       & 0.05/6.65 & 0.03/1.82        \\
 Ofﬁce    & 0.09/3.6      & 0.19/5.42  & 0.21/8.6  & 0.07/1.9       & 0.04/1.95 & 0.07/2.01        \\
 Pumpkin  & 0.08/3.1      & 0.19/3.96  & 0.25/7.2  & 0.08/2.2       & 0.04/2.28 & 0.09/2.26        \\
 Kitchen  & 0.07/3.4      & 0.20/4.94  & 0.26/7.5  & 0.09/2.2       & 0.04/2.21 & 0.09/2.42        \\
 Stairs   & 0.03/2.2      & 0.27/10.6  & 0.28/12.9 & 0.14/3.6       & 0.10/6.48 & 0.14/3.31        \\
\midrule
 Average  & 0.05/2.5      & 0.18/6.55  & 0.22/9.7  & 0.08/3.0       & \textbf{0.05}/3.80 & 0.07/\textbf{2.21}        \\
\midrule
 Kings    & 0.42/0.6      & -          & 0.70/2.92 & 0.33/0.5       & 0.84/1.42 & 0.43/0.87        \\
 Hospital & 0.44/1.0      & -          & 0.97/2.08 & 0.44/0.9       & 1.08/2.41 & 0.46/0.87        \\
 Shop     & 0.12/0.4      & -          & 0.73/4.69 & 0.27/1.6       & 0.59/3.53 & 0.16/0.59        \\
 Church   & 0.19/0.5      & -          & 1.32/3.56 & 0.53/1.6       & 0.63/3.91 & 0.50/1.49        \\
\midrule
 Average  & 0.29/0.63     & -          & 0.92/2.58 & \textbf{0.39}/1.15      & 0.78/2.82 & \textbf{0.39}/\textbf{0.96} \\
\bottomrule
\end{tabular}
}
\end{table*}

\begin{table}[!ht]
\caption{\textbf{(a)} The effect of various level of features on DFNet$_{dm}$ result. Letter F, M, and C denote features extracted from fine, middle, and coarse levels in DFNet.\\\textbf{(b)} Ablation on DFNet (upper part) and \nerfhist in photometric direct matching (lower part). DFM denotes Direct Feature Matching.}
\label{table:6}
\centering
\resizebox{0.8\textwidth}{!}{

\begin{tabular}{cc}

\begin{tabular}{lcccc}
        \multicolumn{5}{l}{\textbf{(a) Featrue level vs. pose error}}\\
        \\
        \toprule
        Feature Level & & & &  DFNet$_{dm}$ (ShopFacade) \\
        \midrule
        F                  & & & &  \textbf{0.15m}, \textbf{0.64°}                   \\
        F+M                & & & &  0.19m, 0.77°                   \\
        F+M+C              & & & &  0.20m, 0.77°                  \\
        \bottomrule
        \end{tabular} & 
\quad \quad \quad
\begin{tabular}{lc}
            \multicolumn{2}{l}{\textbf{(b) Ablation}}\\
            \\
            \toprule
            Method                               & Shop Facade \\
            \midrule
            DFNet w/ $\mathcal{L}_{triplet}^{ori}$ & 1.49m/5.80\degree  \\
            +RVS                                 & 0.86m/4.05\degree  \\
            +$\mathcal{L}_{triplet}$        & 0.72m/2.58\degree  \\
            +DFM (NeRF-W)                       & 0.43m/1.62\degree  \\
            +DFM (NeRF-Hist) & \textbf{0.15m}/\textbf{0.65\degree}  \\
            \midrule
            Direct-PN                            & 1.10m/4.25\degree  \\
            Direct-PN+U                          & 1.41m/6.97\degree  \\
            + NeRF-Hist                          & 0.72m/3.39\degree  \\
            \bottomrule
        \end{tabular}

\end{tabular}
}
\end{table}
\begin{figure}[h]

    \centering
    \begin{tabular}{cc}
    \includegraphics[width=0.4\linewidth]{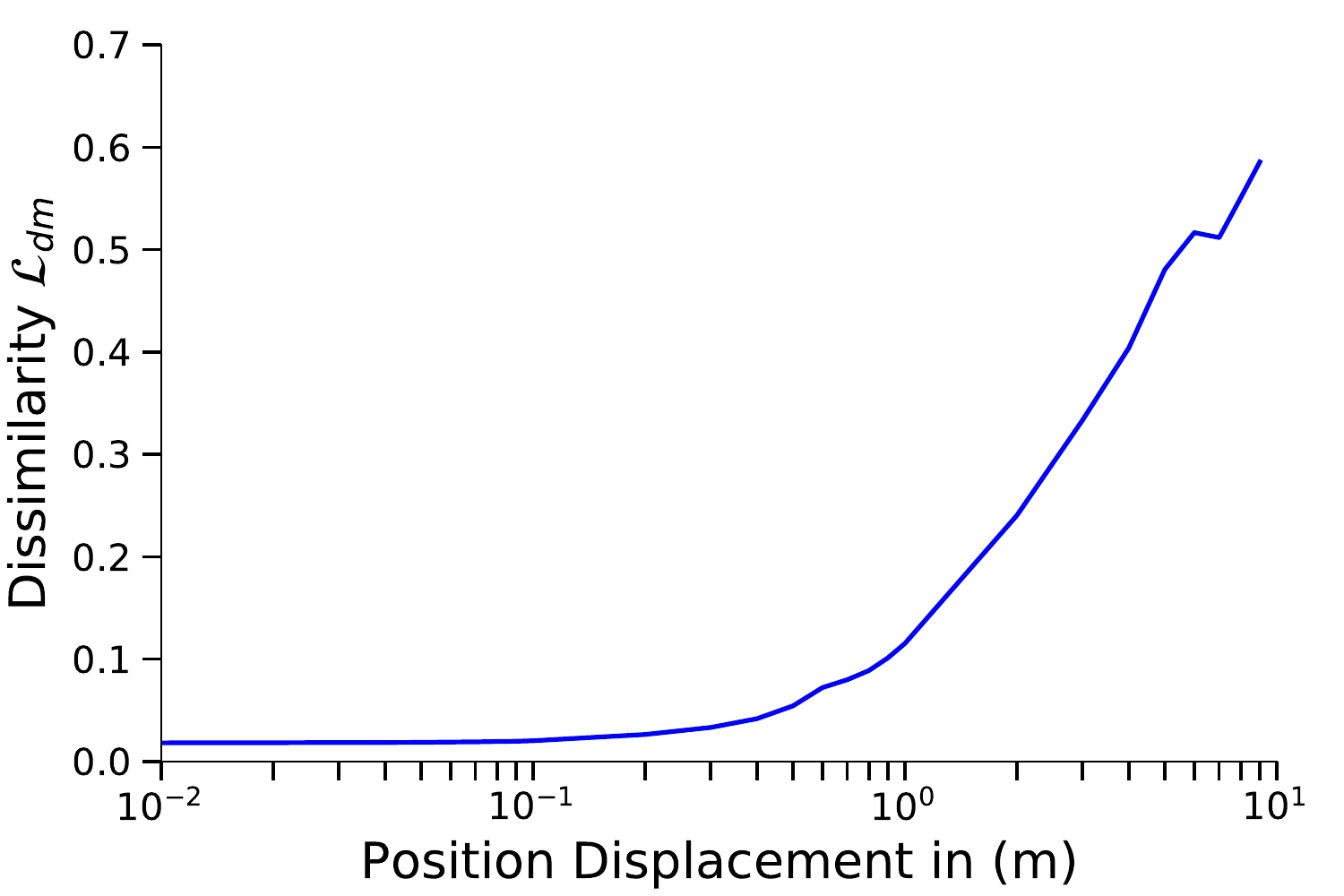}&
    \quad
    \includegraphics[width=0.4\linewidth]{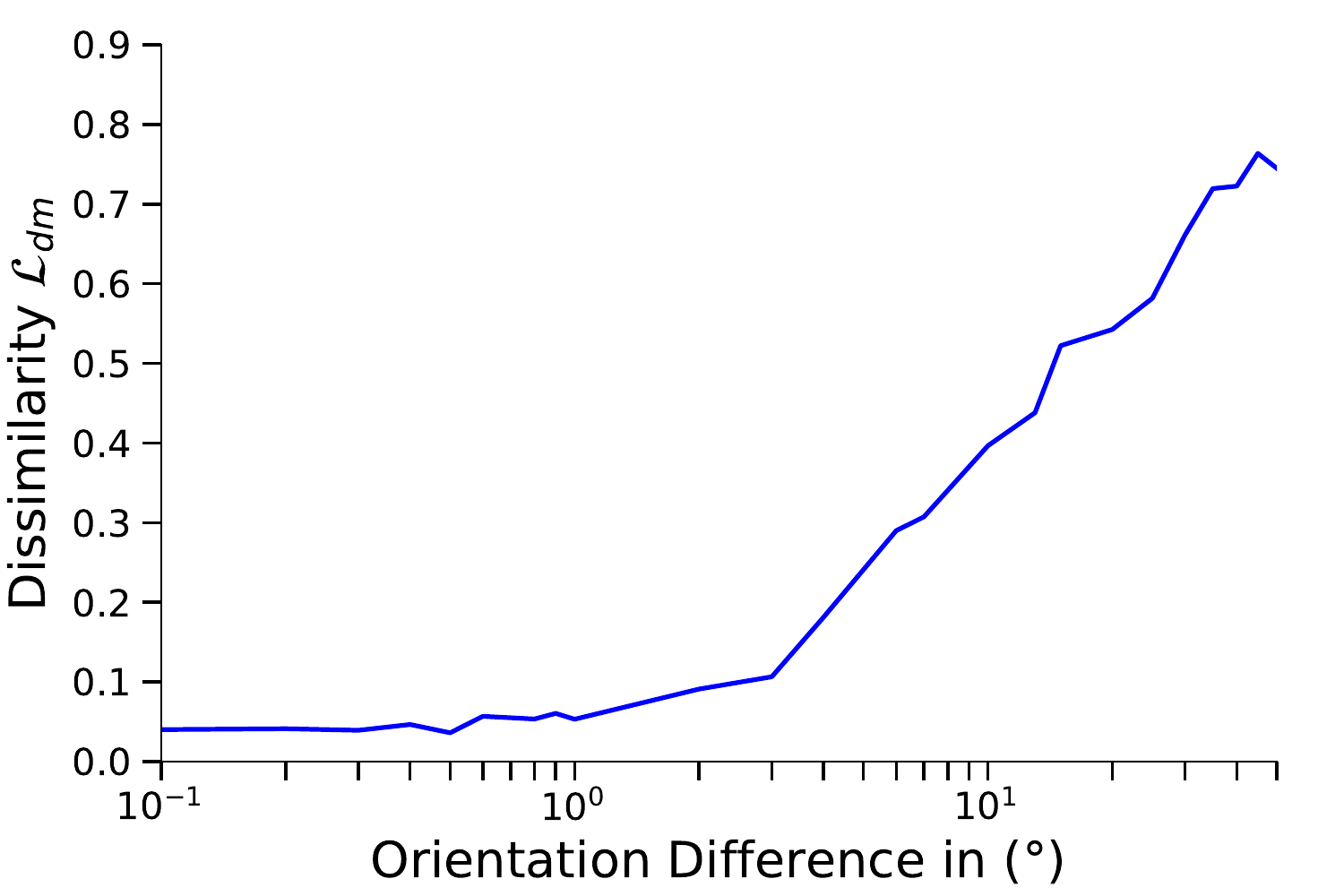}\\

    \end{tabular}
\caption{Pose difference vs. feature dissimilarity. X-axis: camera position (\textbf{left}) and orientation difference (\textbf{right}) between a real image and a rendered image. Y-axis: feature dissimilarity $\mathcal{L}_{dm}$. Our direct feature matching loss $\mathcal{L}_{dm}$ is closely related to pose error, leading to effective training of the APR method.}

\label{fig:DFM_loss}
\end{figure}

\begin{figure}[t]
    \centering
    \begin{tabular}{c}
    \includegraphics[width=0.8\linewidth]{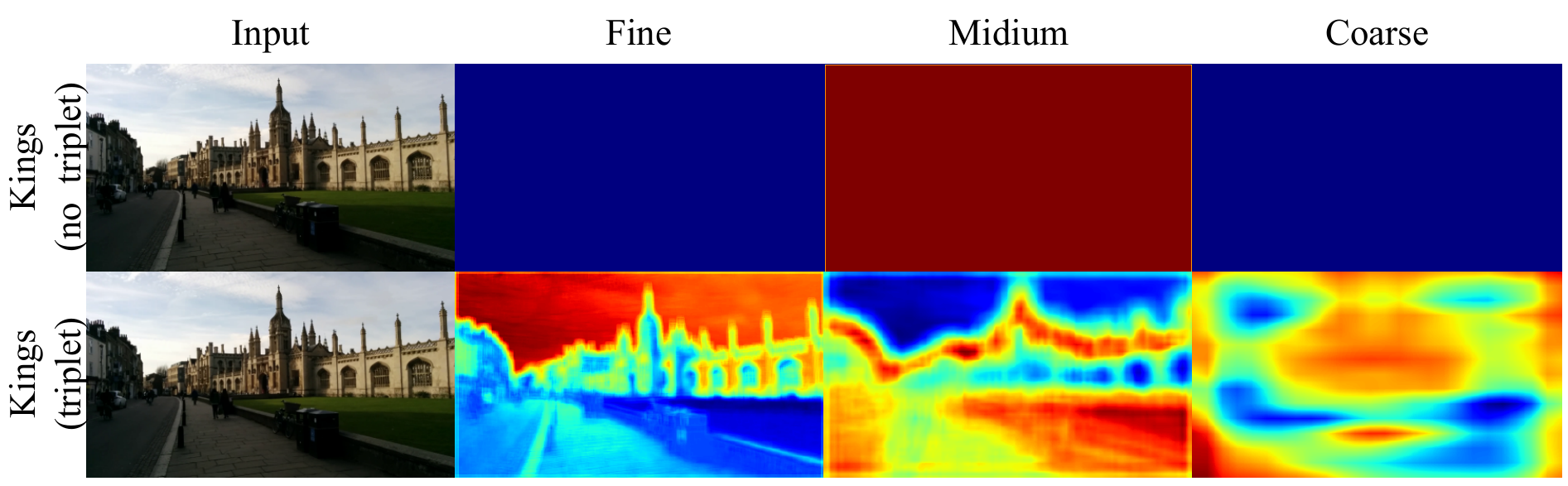}\\
    (a) With and without triplet loss\\
    \includegraphics[width=0.8\linewidth]{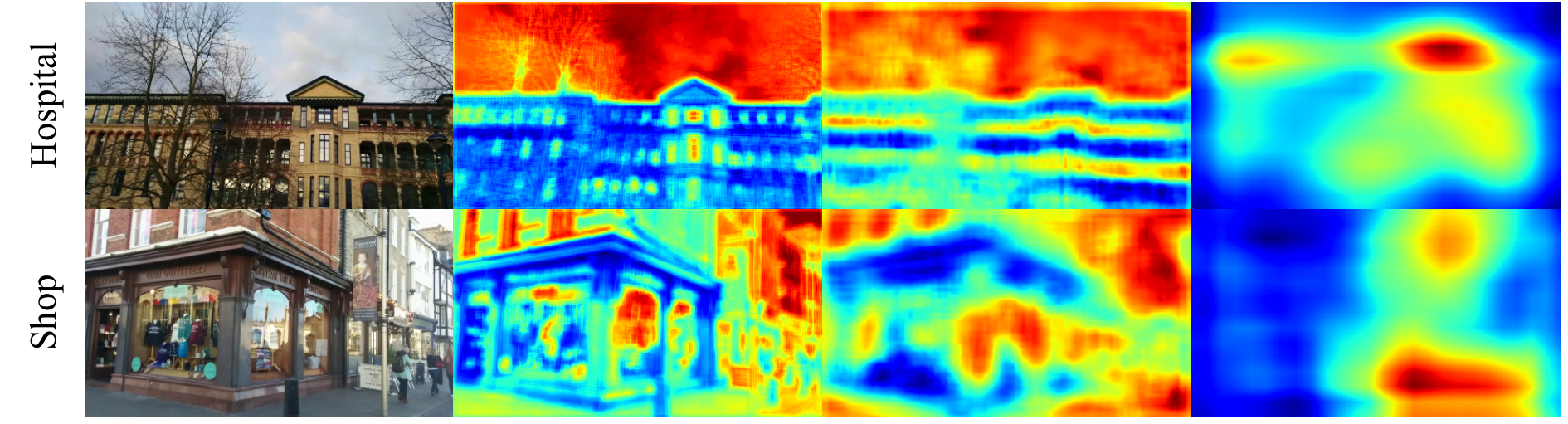}\\
    (b) More feature maps examples when trained with triplet loss
    \end{tabular}
\caption{
\textbf{(a)} Top row: feature collapsing when training DFNet on Kings without using triplet loss. Bottom row: training DFNet with triplet loss can avoid the feature collapsing issue. 
\textbf{(b)} Feature maps of other scenes in Cambridge when training with triplet loss. We show that more refined level features consistently contain more meaningful details and, therefore more beneficial to use for direct feature matching.}
\label{fig:feat_map}

\end{figure}

\begin{table}[h]
\caption{\textbf{Data generation strategy comparison: RVS vs. LENS \cite{Moreau21} on 7-Scenes.} An EfficientNet backbone (as in LENS) is used in DFNet for a fair comparison. Our RVS strategy obtains a comparable results to LENS while using much less training data and rendering in much lower resolution, enabling online training.}
\label{table:7}
\centering
\resizebox{\textwidth}{!}{
\begin{tabular}{lccccccc}
\toprule
Model & \makecell[c]{Backbone\\Top-1 Acc.} & \makecell[c]{Pose Error\\ (m/degree)} & \makecell[c]{Real Data\\Quantity/Epoch} & \makecell[c]{Synthetic Data \\Quantity/Epoch} & \makecell[c]{Synthetic\\Resolution} & \makecell[c]{Rendering\\Cost} & \makecell[c]{Generation\\Mode}\\
\midrule
DFNet(VGG16) & 71.59\% & 0.12/3.71 & 10-20\% & 10-20\% & Low & Cheap &Online\\
DFNet(EB0)   & 76.3\%  & 0.08/3.47 & 10-20\% & 10-20\% & Low & Cheap &Online\\
LENS(EB3)    & 81.1\%  & 0.08/3.00 & 71\%-100\% & 710\%-1000\% & High & Expensive & Offline\\
\bottomrule
\end{tabular}
}

\end{table}

\subsection{Comparison to Sequential APR and 3D Approaches}
\cref{table:3} compares our method to other types of relocalization approaches, such as several state-of-the-art sequential-based APR approaches and 3D structure-based method Active Search \cite{Sattler12}. We notice that our DFNet$_{dm}$ outperforms most sequential-based APR methods except the translation error of VLocNet \cite{Valada18} on 7-Scenes in terms of the scene average performance. However, we still achieve superior accuracy than VLocNet in 7 out of 11 scenes. For the first time, the performance of single-image APR is comparable to 3D-structure methods. Our DFNet$_{dm}$ is slightly more accurate than Active Search \cite{Sattler12} in average rotation error of 7-scenes. However, our method is still slightly behind in terms of translation error and Cambridge errors although by smaller margins.

\subsection{Ablation Study} \label{sec:additional_insights}
\subsubsection{Effectiveness of Direct Feature Matching} \label{sec: DFM}
We run a toy example of direct feature matching on Shop Facade using finest features and combinations of multi-level features, as in  \cref{table:6}(a). We discover that finer-level features are more helpful for direct matching. We believe this to be due to their capability to preserve high frequency details and sharper contents, as shown in (\cref{fig:feat_map}(b)). This explains why we only use the finest feature in the feature-metric direct matching implementation.
Furthermore, \cref{fig:DFM_loss} shows how the direct matching loss $\mathcal{L}_{dm}$ successfully correlates the pose differences to the feature similarity between real images and rendered images.

\subsubsection{Features Collapse}
We demonstrate the difference when training DFNet's feature extractor with and without triplet loss in \cref{fig:feat_map}(a). We replace our triplet loss with a mean square error (MSE) loss for the without triplet loss case. Intuitively, losses that only minimize positive sample distances such as MSE, $L_2$, or $L_1$ losses may lead to feature collapsing \cite{Chen21explore} since the feature extraction blocks in DFNet are likely to learn to cheat. On the other hand, using triplet loss supervised with additional negative samples works well for extracting dense domain invariant features.

\subsubsection{Summary of Ablation}
We break down our design decisions to show how each component contributes to the pose regression accuracy in \cref{table:6}(b). We start with training an DFNet model using with standard triplet loss without mining. 
The performance improves noticeably when we add the RVS. We also see around 16\%/36\% gain in translation and rotation errors when adding the customized triplet loss $\mathcal{L}_{triplet}$ . We then validate our DFNet$_{dm}$'s direct feature matching (DFM), which further reduces error significantly. The DFM approach with \nerfhist outperforms the NeRF-W one, which validates the effectiveness of our histogram embedding design. Finally, we attempt to train a Direct-PN+U model with our \nerfhist modification. Our results show that the photometric direct matching-based method that can benefit from our new NVS method, though the pose estimation accuracy is worse than our feature-metric direct matching method.

\subsubsection{Effectiveness of RVS} \label{sec:effectiveness_of_rvs}
\cref{table:7} shows a comparison between our online RVS strategy with another peer work LENS \cite{Moreau21} that uses NeRF data generation for APR training. Although both data generation methods effectively improve APR performance, our RVS strategy is a much cheaper alternative requiring lower rendering resolution (80x60 vs. 320x240 \cite{Moreau21}) and fewer data. We are able to reach similar performance with LENS when we replace our VGG16 backbone with an EfficientNet-B0 \cite{Tan19}, which proves that a simpler data generation strategy could also effectively improves APR methods.


\section{Conclusion} \label{sec:conclusion}
In summary, we introduce an Absolute Pose Regression (APR) pipeline for camera re-localization. 
Specifically: 1) we propose a \nerfhist to compensate dramatic exposure variance in large scale scene with challenging exposure conditions. 
The \nerfhist, serving as a novel view renderer, enables a direct matching training scheme;
2) we explore a direct matching scheme in feature space, leading to a more robust performance than the photometric approach, and address a domain gap issue that arises when matching real images with synthetic images via a contrastive learning scheme;
3) we devise an efficient data generation strategy, which proposes pseudo training poses around existing training trajectories, leading to better generalization capability to unseen data.
As a result, our method achieves a state-of-the-art accuracy by outperforming existing single-image APR methods by as much as 56\%, comparable to 3D structure-based methods.

\subsubsection{Acknowledgments}
The authors thank Michael Hobley, Theo Costain, Lixiong Chen, and Kejie Li for their thoughtful comments. Shuai Chen was supported by gift funding from Huawei.






\clearpage
%
%
\bibliographystyle{splncs04}
\bibliography{main}

\clearpage
\section{Supplementary}
\subsection{Implementation}
\subsubsection{Histogram-assisted NeRF}
Here, we provide more implementation details of our methods. As we mentioned in Section 3.3 of the main paper, our histogram-assisted NeRF model renders at a speed of 12.2 fps (benchmarked by a 3080Ti GPU) to achieve online RVS training. In order to achieve a balanced trade-off between speed and quality, we choose to render small images with a shorter side of 60 pixels. In addition, we set the NeRF model architecture to 64 coarse and 64 fine sampling with an MLP width of 128.

\subsubsection{DFNet} Our DFNet takes an image input with a shorter side of 240 pixels. For feature extractor module $\mathcal{G}$ of DFNet, features are fed through a Conv-Relu-Conv-Batch Norm architecture with 64 kernels and 128 output channels. The DFNet is trained with a batch size of 4 or 8, depending on the GPU's memory. We implement an early stopping strategy with a patient value of 200 and schedule the learning rate decay of 0.95 when validation loss plateaus for every 50 epochs. For every $N=20$ epochs, we will randomly generate the same amount of views as the training sample size using RVS.

\subsubsection{Direct Feature Matching} To train the DFNet$_{dm}$ model with feature-metric direct matching using unlabeled data, we set the batch size to 1 and the learning rate to $1 \times 10^{-5}$ with the same early stopping strategy mentioned above. We discover that only low-level features (i.e., features from the first blocks of VGG) are needed to achieve the best performance, which we discussed earlier in Section 4.5 of the main paper.
\begin{figure}[ht]
  \centering
    \begin{tabular}{cccc}
      \includegraphics[width=0.25\linewidth]{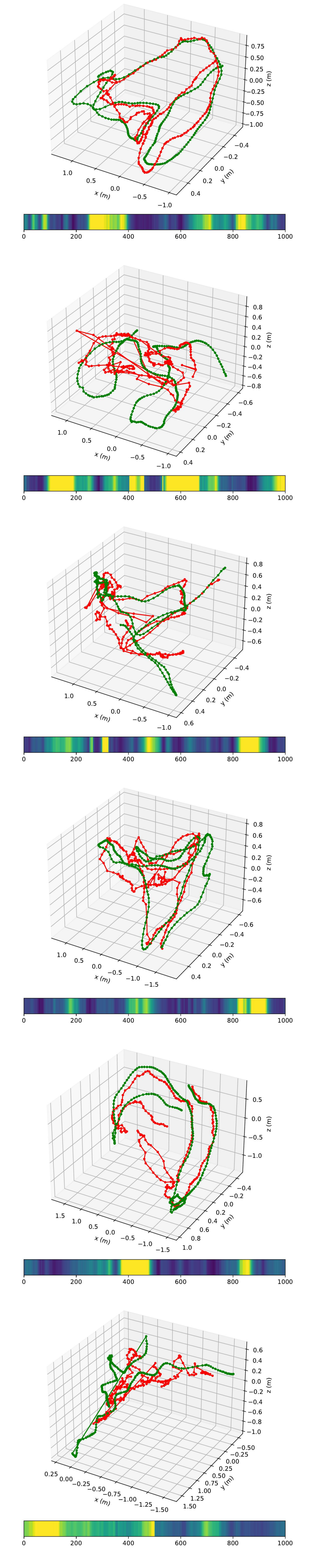} &
      \includegraphics[width=0.25\linewidth]{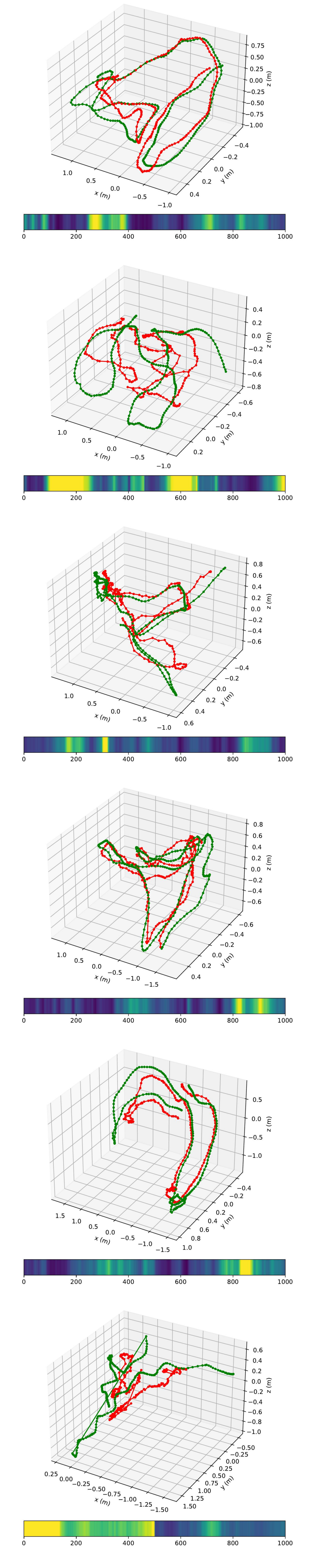}&
      \includegraphics[width=0.25\linewidth]{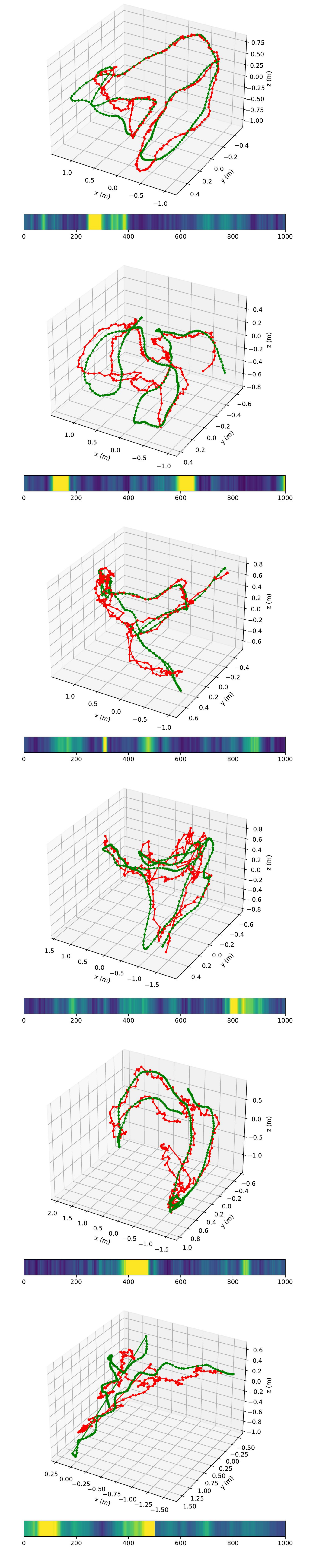}&
      \includegraphics[width=0.25\linewidth]{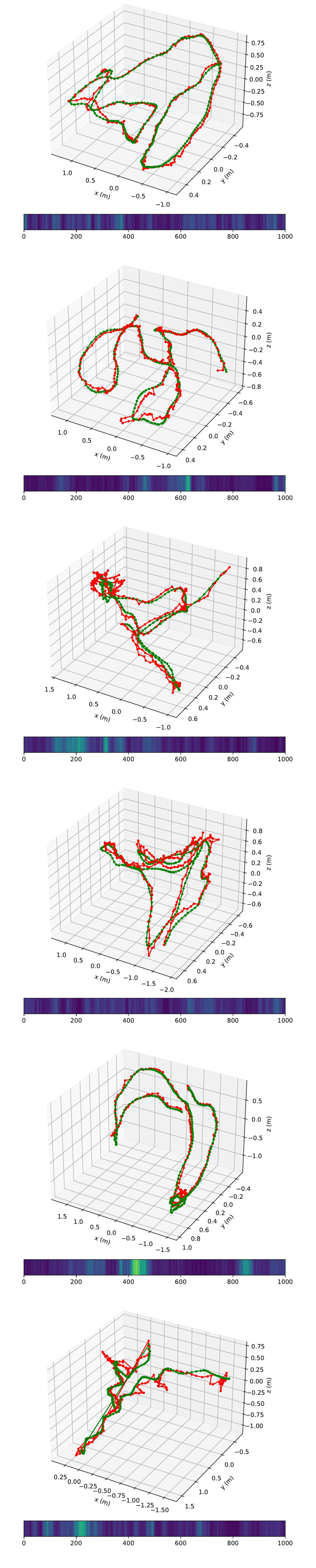}\\
      (a) PoseNet\cite{Kendall15,Kendall16,Kendall17} & (b) MapNet+PGO\cite{Brahmbhatt18} & (c) Direct-PN+U\cite{chen21} & (d) DFNet$_{dm}$
    \end{tabular}
\caption{Qualitative comparison on the 7-Scenes dataset. The 3D plots show the camera positions, \textcolor{green}{green} for ground truth and \textcolor{red}{red} for predictions. The bottom color bar represents rotational errors for each subplot, where yellow means large error and blue means small error for each test sequence. Sequence names from top to bottom are: Chess-seq-03, Fire-seq-04, Office-seq-07, Kitchen-seq-06, Kitchen-seq-12, Stairs-all.}
\label{fig:7scenes_qualitative}
\end{figure}
\begin{figure}[ht]
    \centering
    \includegraphics[width=1.0\linewidth]{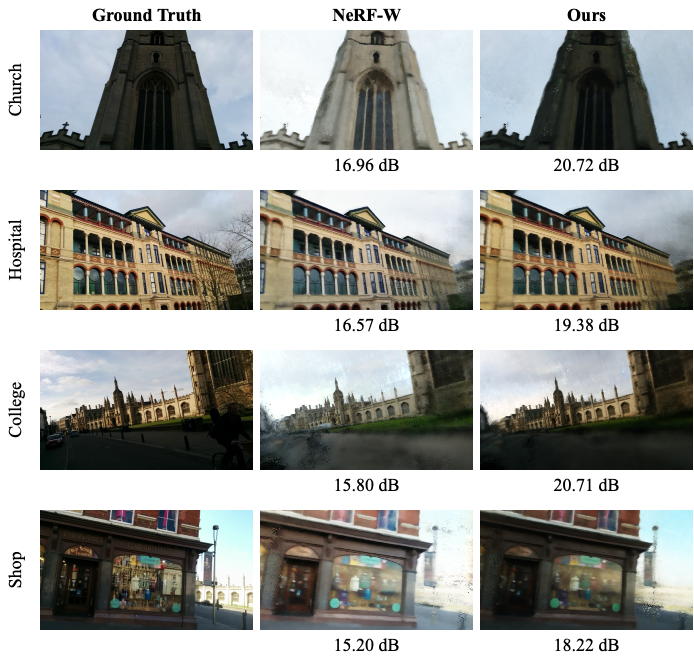}
\caption{A visual comparison between NeRF-W and our histogram-assisted NeRF on the testing sequences of Cambridge Landmarks dataset. The corresponding scene's test PSNR is displayed at the bottom of each sub-figure.}
\label{fig:NeRF-hist_supp}
\end{figure}
\subsection{Visualization}
\subsubsection{Qualitative Comparison on 7-Scenes}
We show a selection of qualitative comparisons on the 7-Scenes dataset with several baseline APR methods \cite{Kendall17,Brahmbhatt18,chen21} in \cref{fig:7scenes_qualitative}. We also encourage our readers to check out our supplementary video, in which we rendered views of the predicted pose using NeRF synthesis.

\subsubsection{NeRF-W vs. Histogram-assisted NeRF}
In real-life camera localization applications, since training and testing data are likely to be taken from different sequences, camera exposures, or time of the day, our histogram-assisted NeRF would be more helpful to render accurate appearcance \cref{fig:NeRF-hist_supp}. We experimentally found our histogram-assisted NeRF is helpful in both photometric matching and feature-metric matching approaches.

\subsection{Additional Discussion}
\subsubsection{Photometric Distortion}
As discussed in section 3.2 of the main paper, photometric matching relies on RGB-wise differences between the query and rendered images. However, if those images appear in different lighting/exposure conditions, the photometric loss will fail due to large RGB-wise differences even under the same camera pose. Previous photometric matching approaches, such as Direct-PN+U, perform worse in pose estimation when using unlabeled data with large appearance variations from the training sequences (refer to the lower part of main paper Table 4b). We observed that such degradation is consistent in other outdoor scenes.

\subsubsection{Two Properties of Domain Invariant Features}
We had two clear goals for designing our robust feature extractor: (1) We want the extracted features to be sensitive to pose changes. (2) We want the features to be indistinguishable between real and rendered image features from the same pose (Close the Domain Gap). Our first goal is achieved by the $L_2$ pose loss supervision, which ensures the features are closely related to the pose regression task. We specifically design the Feature Extractor to share the backbone with the Pose Module (see main paper Fig 2a). Although the deeper layer features may lose semantic meaning, we observe that those features can respond to pose changes.

The triplet loss is primarily designed to achieve the second goal without feature collapse in the training process. We previously tried to force real and rendered image features to be the same by using MSE/$L_2$ losses, leading to feature collapse (main paper Fig 6a). This is because the pink layers in main paper Fig 2a, despite being shallow, are likely to learn to cheat since those layers are not supervised by other meaningful losses. Thus, we introduce the triplet loss to prevent features collapsing. We experimentally find that the proposed in-triplet mining adds extra robustness to both feature extraction and pose regression and leads to better APR performance overall. Such observation could hint that removing the domain gap benefits APR training when using extra randomly generated synthetic training data.
\end{document}